\pdfoutput=1

\documentclass[11pt]{article}

\usepackage{booktabs}
\usepackage{caption}
\usepackage{graphicx}
\usepackage{multirow}
\usepackage{tabularx}
\usepackage{hyperref}

\usepackage[most]{tcolorbox} 
\usepackage{fontawesome5} 
\usepackage{enumitem} 

\usepackage[final]{coling}

\usepackage{times}
\usepackage{latexsym}

\usepackage[T1]{fontenc}

\usepackage[utf8]{inputenc}

\usepackage{microtype}

\usepackage{inconsolata}

\usepackage{graphicx}

\title{Aligning Large Language Models to Low-Resource Languages through LLM-Based Selective Translation: A Systematic Study}

\author{Rakesh Paul, Anusha Kamath, Kanishk Singla, Raviraj Joshi\\ \textbf{ Utkarsh Vaidya, Sanjay Singh Chauhan, Niranjan Wartikar} \\
        NVIDIA \\
        \texttt{\{rapaul, anushak, kanishks, ravirajj, uvaidya} \\ \texttt{schauhan, nwartikar\}@nvidia.com}}

\begin{document}
\maketitle
\begin{abstract}
Multilingual large language models (LLMs) often demonstrate a performance gap between English and non-English languages, particularly in low-resource settings. Aligning these models to low-resource languages is essential yet challenging due to limited high-quality data. While English alignment datasets are readily available, curating equivalent data in other languages is expensive and time-consuming. A common workaround is to translate existing English alignment data; however, standard translation techniques often fail to preserve critical elements such as code, mathematical expressions, and structured formats like JSON.
In this work, we investigate LLM-based selective translation, a technique that selectively translates only the translatable parts of a text while preserving non-translatable content and sentence structure. We conduct a systematic study to explore key questions around this approach, including its effectiveness compared to vanilla translation, the importance of filtering noisy outputs, and the benefits of mixing translated samples with original English data during alignment. Our experiments focus on the low-resource Indic language Hindi and compare translations generated by Google Cloud Platform (GCP) and Llama-3.1-405B. The results highlight the promise of selective translation as a practical and effective method for improving multilingual alignment in LLMs.
\end{abstract}

\section{Introduction}
Large Language Models (LLMs) have achieved remarkable success across various natural language processing tasks, largely driven by vast amounts of high-quality English data \cite{anil2023palm,achiam2023gpt,bercovich2025llama}. However, a significant performance disparity persists when these models are applied to non-English languages, especially those designated as low-resource \cite{joshi2020state,jadhav2024limitations}. Bridging this gap is critical for equitable AI development and broader global applicability. The primary impediment to aligning LLMs with low-resource languages lies in the scarcity of high-quality, diverse, and representative training data \cite{cahyawijaya2024llms}. While comprehensive English alignment datasets are abundant, the creation of analogous resources in other languages is often expensive and time-consuming.

Current approaches for low-resource adaptation of language models include continued pre-training using low-resource data, which helps in familiarizing the model with the target language's unique linguistic characteristics \cite{joshi2024adapting}. Another prominent method is alignment using low-resource supervised fine-tuning (SFT) and preference tuning, where models are trained on specific downstream tasks and human feedback to better adhere to user intent and safety guidelines in the low-resource language \cite{li2024x,toraman2024adapting}. The data for these alignment processes is typically curated using synthetic data generation methods, with translation of high-resource data being a common strategy \cite{qin2024multilingual}. Other probable methods include cross-lingual transfer learning, where knowledge from high-resource languages is transferred to low-resource languages, and techniques like zero-shot or few-shot learning, which leverage the model's inherent generalization capabilities to perform tasks with minimal or no explicit low-resource data \cite{cahyawijaya2024llms,lai2024llms}. Additionally, active-learning, self-training and semi-supervised learning methods, which utilize unlabeled low-resource data, are also being explored \cite{kholodna2024llms}.

\begin{figure*}[h]  
    \centering
    \includegraphics[scale=0.1]{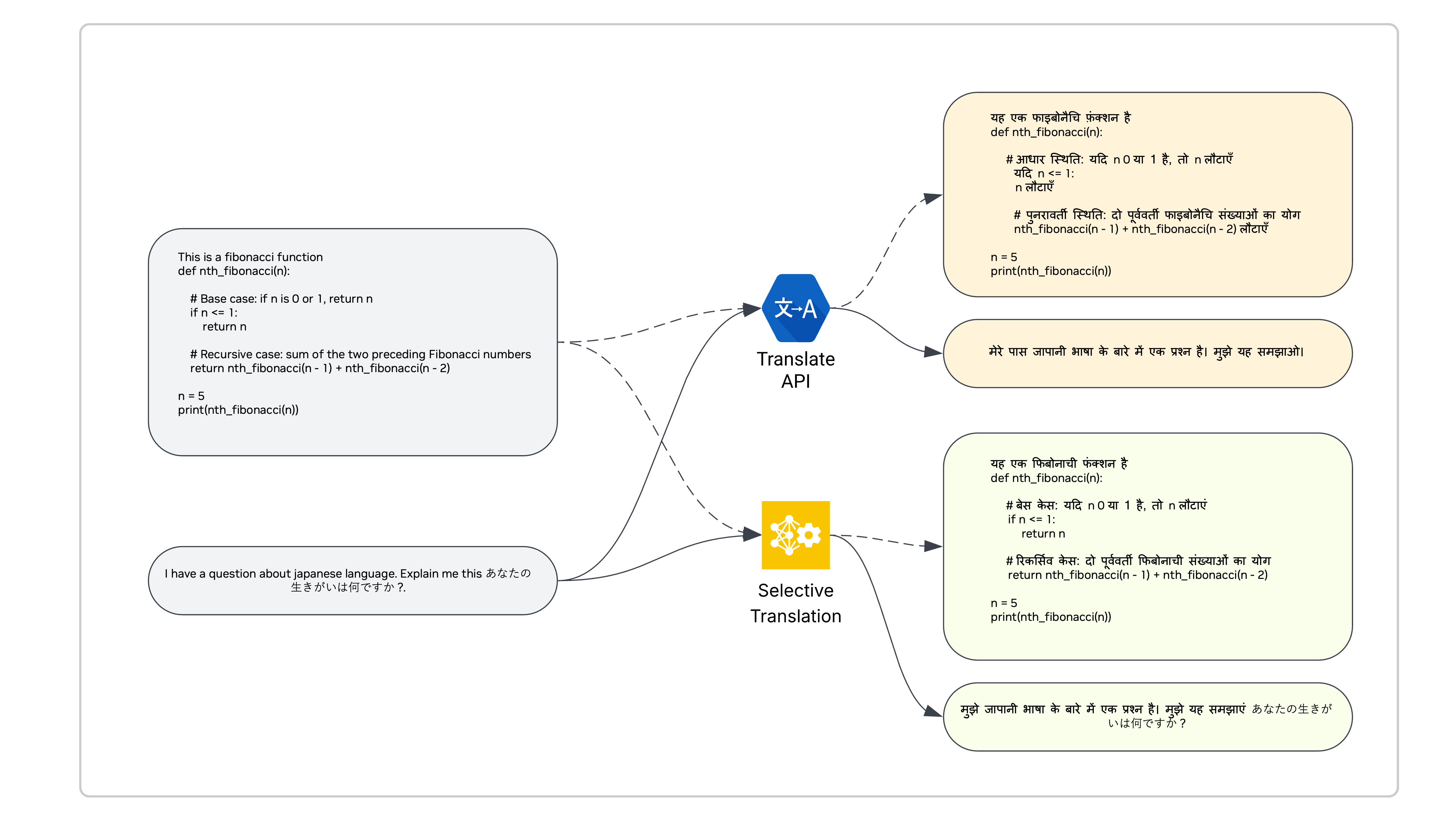}  
    \caption{English to Hindi translation examples using LLM-based selective translation and vanilla GCP translation.}
    \label{fig:selective_translation_examples}
\end{figure*}

A common and seemingly straightforward approach to address this data scarcity is to translate existing high-resource (e.g., English) alignment datasets into the target low-resource language. Nevertheless, conventional machine translation techniques frequently fall short. They often struggle to accurately preserve crucial non-translatable elements such as code snippets, complex mathematical expressions, and structured formats like JSON, leading to corrupted or functionally incorrect data. This issue severely limits the utility of conventionally translated datasets for robust LLM alignment, particularly for tasks requiring precise understanding of structured or logical content.

To address these limitations, we systematically investigate LLM-based selective translation, an approach that intelligently translates only the linguistically appropriate portions of a prompt while preserving non-translatable content and maintaining overall sentence structure. This method leverages the reasoning capabilities of LLMs to distinguish between translatable and non-translatable segments, offering a more faithful and usable translation for alignment purposes. The distinct advantages of LLM-based selective translation over vanilla machine translation are demonstrated in Figure \ref{fig:selective_translation_examples}.

In this study, we explore three key research questions:
\begin{itemize}
    \item How does LLM-based selective translation of alignment data compare to conventional (vanilla) translation methods, such as Google Cloud Platform (GCP), on the performance of the aligned model?
    \item What is the optimal strategy for mixing original English alignment data with selectively translated target language data? Can translated data alone achieve effective alignment, or is the inclusion of English data indispensable?
    \item What is the impact of filtering noisy or erroneous outputs generated during the selective translation process?
\end{itemize}

We train the Nemotron-4-Mini-Hindi-4B-Base\footnote{\url{https://huggingface.co/nvidia/Nemotron-4-Mini-Hindi-4B-Base}} \cite{joshi2024adapting} model on the LLM-translated and GCP-translated datasets and compare the performance on the downstream tasks like MTBench, IFEval, and GSM8K in Hindi. Our experiments focus on Hindi, a widely spoken yet low-resource Indic language. We compare translation quality and alignment effectiveness across outputs generated by GCP and Llama-3.1-405B, a powerful open-source LLM. Through this comprehensive analysis, we demonstrate that LLM-based selective translation offers a practical and robust solution for multilingual alignment, substantially improving the performance of LLMs in low-resource settings and moving toward more linguistically inclusive AI systems. Throughout this study, "LLM translation" specifically refers to translations produced by Llama-3.1-405B. 

\begin{figure*}[h]  
    \centering
    \includegraphics[scale=0.6]{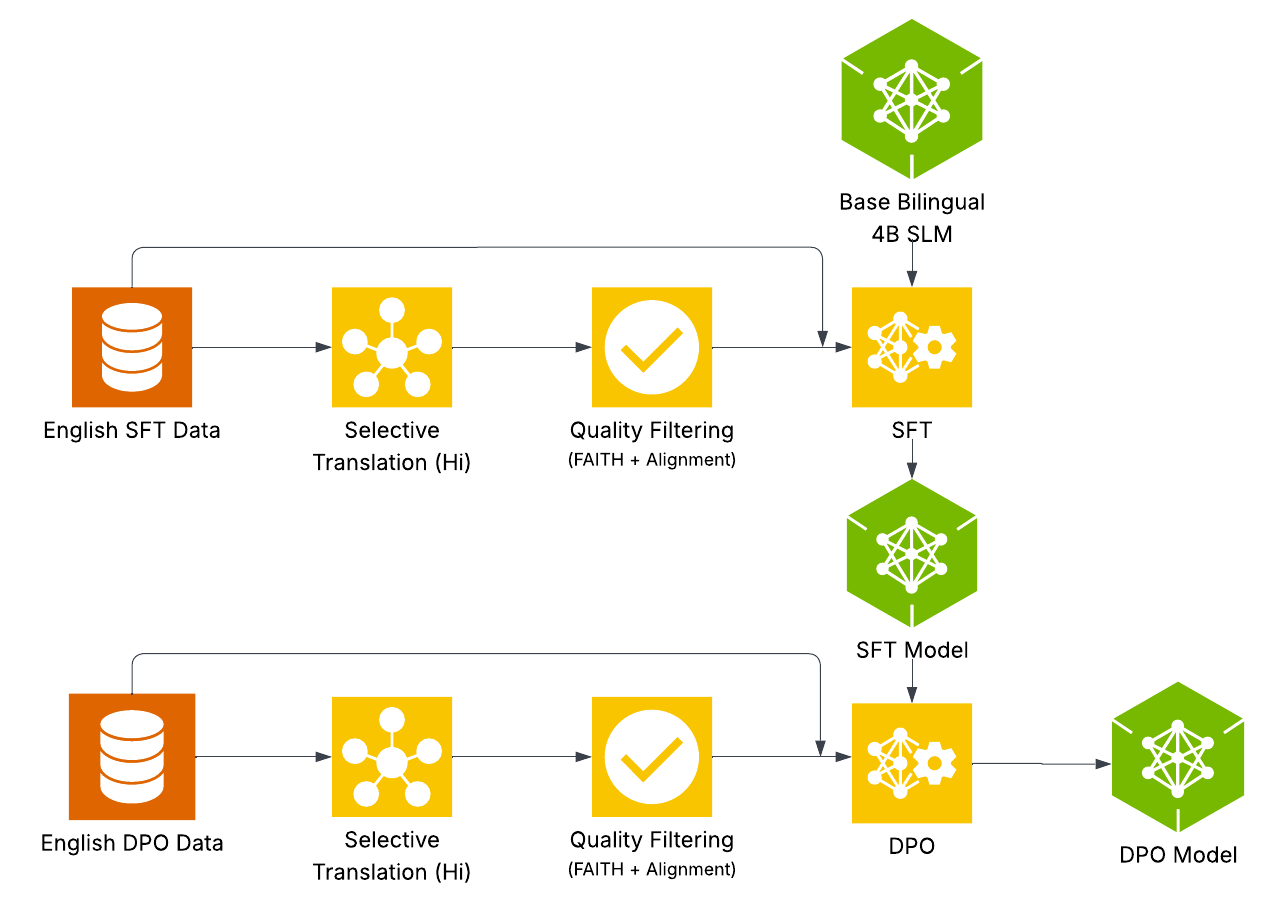}  
    \caption{Overall training pipeline comprising translation, filtering, SFT, and DPO stages.}
    \label{fig:selective_translation}
\end{figure*}

\section{Related Work}
Large Language Models (LLMs) have demonstrated impressive capabilities, primarily due to extensive training on high-resource language data, leading to a performance disparity in low-resource languages (LRLs). To address this, various adaptation strategies have been explored, including continued pre-training on limited, authentic, or synthetically generated LRL corpora \cite{joshi2024adapting}. \cite{ogueji2021small} indicates that even modest amounts of LRL exposure can yield significant improvements, while \cite{hangya2022improving} delves into specific techniques within multilingual frameworks. Furthermore, the development of dedicated open-source LLMs for languages like Hindi, as exemplified by \cite{choudhury2025llama}, underscores the importance of tailored training on relevant LRL datasets.

Multilingual LLMs (MLLMs) offer a promising avenue for addressing LRL challenges, with their inherent capacity for zero-shot or few-shot cross-lingual transfer, including in-context learning abilities for LRLs \cite{cahyawijaya2024llms}. However, this multilinguality does not guarantee uniform performance across all languages; empirical studies reveal significant disparities, with high-resource languages often outperforming LRLs \cite{hasan2024large}. In some cases, multilinguality can even pose a "curse" where LRL performance is hindered due to disproportionate resource allocation to high-resource languages \cite{chang2024multilinguality}. To counteract these limitations and facilitate more effective knowledge generalization, methods like cross-lingual optimization have been proposed to enhance language transfer from high-resource to low-resource settings \cite{lee2025cross}.

Beyond foundational training, instruction tuning has become crucial for aligning LLMs with human intent. In a multilingual context, this extends to cross-lingual instruction following and explicit alignment mechanisms. \cite{cahyawijaya2023instructalign} demonstrates the effectiveness of continual cross-lingual instruction tuning for aligning languages, while \cite{tanwar2023multilingual} emphasizes the role of alignment in boosting cross-lingual in-context learning. \cite{ahuja2024sphinx} explores sample-efficient multilingual instruction fine-tuning via guided prompting. A broader perspective on enhancing multilingual capabilities and alignment strategies is provided by \cite{zhao2024lens}, collectively emphasizing the move towards task-oriented instruction following and robust cross-lingual alignment.

While existing research extensively covers continued pre-training, diverse fine-tuning strategies, and instruction-based alignment for LRLs, a systematic exploration of leveraging LLM-based selective translation as a primary alignment mechanism remains largely underexplored.

\section{Methodology}

\subsection{Model Alignment}
The alignment of multilingual large language models to low-resource languages is a critical step in bridging the performance gap observed between high-resource and low-resource settings. In this work, we employ a two-stage alignment process, Supervised Fine-Tuning (SFT) followed by Direct Preference Optimization (DPO). Both stages are designed to leverage the strengths of our selectively translated Hindi corpus alongside the original English corpus, ensuring a robust and multilingual alignment.

\begin{itemize}
    \item \textbf{Supervised Fine-Tuning (SFT):}
    During SFT, the Nemotron-4-Mini-Hindi-4B-Base model is fine-tuned on a dataset of high-quality instruction-response pairs. The primary goal of this stage is to teach the model to follow instructions and generate coherent, relevant responses.

For SFT, we utilize a mixed corpus comprising both the original English alignment dataset and its selectively translated Hindi counterpart. We use an English SFT corpus with approximately 200k examples, comprising various tasks as outlined in \cite{adler2024nemotron}. This mixed approach is crucial for retaining the model's English capabilities, adapting it to Hindi's linguistic nuances, and ensuring correct handling of non-translatable content like code and mathematical expressions across both languages.

The SFT process is performed using a standard cross-entropy loss function, optimizing the model's parameters to predict the correct response given an instruction. This stage is followed up by the subsequent preference-based optimization.

    \item \textbf{Direct Preference Optimization (DPO):}
    Following SFT, we apply DPO to further refine the model's alignment with human preferences and improve its ability to generate helpful and harmless responses. DPO is a reinforcement learning from human feedback (RLHF) alternative that directly optimizes a policy to align with human preferences without requiring a separate reward model.

For the DPO stage, we construct preference datasets consisting of pairs of responses (one preferred, one rejected) for a given prompt. Similar to SFT, these preference datasets are also derived from a combination of original English preference data and selectively translated Hindi preference data.
The DPO algorithm directly optimizes the policy by maximizing the log-probability of preferred responses and minimizing the log-probability of dispreferred responses, effectively aligning the model with the implicit reward signal encoded in human preferences. This final stage fine-tunes the model to produce responses that are not only accurate but also preferred by users in both high-resource and low-resource language settings.
\end{itemize}
Across both SFT and DPO stages, we utilized 64 A100 GPUs. The learning rate was set with a maximum of 5e-6 and a minimum of 9e-7, employing a cosine annealing schedule. The batch size for SFT and DPO was set to 1024 and 512, respectively. The models were trained for 2 epochs in both stages using Nemo Aligner \cite{shen2024nemo}. The overall training process is highlighted in Figure \ref{fig:selective_translation}.

\subsection{Selective Translation}
Conventional translation of SFT or DPO data typically involves translating entire text segments without specific consideration for their inherent structure or non-linguistic components. This approach, while broadly useful for general text, often faces significant limitations when applied to specialized datasets crucial for LLM alignment. Specifically, it struggles to accurately preserve critical elements such as programming code, URLs, file paths, email addresses, highly formatted data (e.g., tables, lists), examples where direct translation would alter their original meaning or usefulness, special characters, mathematical symbols, technical abbreviations, and HTML/XML tags. This can lead to corrupted data that loses its functional integrity, rendering it less effective or even counterproductive for training LLMs on tasks requiring precise understanding of such structured or technical content.

Selective translation using LLMs is a technique where a Large Language Model is specifically instructed to translate only the linguistically adaptable portions of a given text, while meticulously preserving certain non-translatable elements. This approach, unlike conventional translation, prevents the corruption of critical content such as programming code, URLs, file paths, email addresses, highly formatted data (tables, lists), examples where meaning would be lost, special characters, mathematical symbols, technical abbreviations, and HTML/XML tags. By following precise rules, which are specified as part of the prompt, the LLM intelligently identifies and skips these specific segments, ensuring they remain unchanged in the output. Furthermore, unlike typical machine translation solutions that translate line by line, the selective translation approach used in this work processes the entire prompt or response at once, thereby maintaining crucial inter-sentence coherence. The goal is to produce naturally flowing translated sentences that maintain their original structure and accurately retain all functional or context-sensitive non-linguistic information. This enables high-fidelity multilingual data generation, which is especially crucial for technical or structured content. The exact prompt used for selective translation is presented in the Appendix \ref{sec:appendix}.

\subsection{Quality Filtering}
The process of generating translated data, particularly through LLM-based approaches, inherently introduces the risk of noisy or erroneous translations. Such noise can significantly impede the downstream alignment process of LLMs, potentially leading to the propagation of errors, reduced model performance, and a suboptimal learning experience for the target language. Therefore, a robust quality filtering mechanism is crucial to ensure that only high-fidelity translated samples are used for SFT or DPO.

To address this, we implement a FAITH-based filtering mechanism utilizing LLMs. FAITH considers five crucial aspects for comparing original and translated samples: Fluency, Accuracy, Idiomaticity, Terminology, and Handling of Format. This approach leverages the generative and evaluative capabilities of a separate LLM to assess the quality of the translated outputs against the original source sentences. The LLM acts as an automated evaluator, scoring translations across these critical dimensions. The prompt used for this evaluation is presented in the Appendix \ref{sec:appendix}. The Llama-3.1-Nemotron-70B-Instruct model was used for FAITH-based filtering; we only retain examples that receive full scores of 5 across all the parameters from the judge LLM.

Following the FAITH-based filtering, we apply an additional layer of alignment-based filtering. This process specifically measures how well the translated prompt and its corresponding translated response align with each other post-translation. It evaluates the logical consistency and coherence between the translated query and its response, using metrics such as Helpfulness, Correctness, Coherence, Complexity, and Verbosity. Each metric is scored on a scale of 1 to 5, ensuring that the retained data not only exhibits high translation fidelity but also maintains the intended relationship and quality between the prompt and response, further refining the training corpus for optimal alignment.


\subsection{Safety Data Considerations}
The SFT and DPO datasets incorporate unsafe samples. These samples refer to queries and responses that contain harmful, biased, or inappropriate content, and are crucial for training the model to appropriately refuse or handle such questions. While contemporary LLMs often inherently refuse to translate unsafe content, traditional translation solutions like GCP typically translate these queries without refusal.

\begin{figure*}[h]  
    \centering
    \includegraphics[scale=0.6]{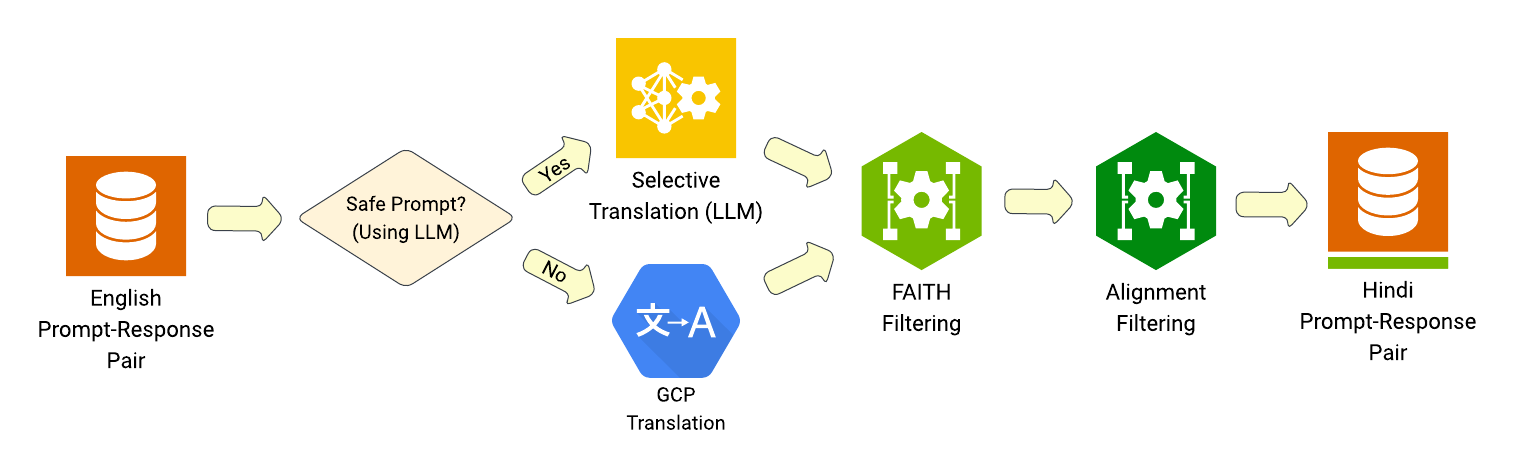}  
    \caption{Hybrid approach for selective translation-based data curation pipeline with safety considerations. The unsafe queries contain harmful, biased, or inappropriate content that LLMs typically decline to translate.}
    \label{fig:safety_considerations}
\end{figure*}

Therefore, we adopt a hybrid approach for safety-critical data. Initially, a Safety-Guard LLM, specifically Llama-Nemotron-Safety-Guard-v2\footnote{\url{https://huggingface.co/nvidia/llama-3.1-nemoguard-8b-content-safety}}, is employed to classify prompts and responses as either safe or unsafe. All identified unsafe samples are then consistently translated using GCP. This ensures that even within the LLM-translated data, unsafe examples are processed via GCP, allowing the model to learn refusal behaviors from these translated unsafe queries. It is important to note that unsafe samples constitute approximately 5\% of our total dataset. 
The full pipeline for the hybrid approach is shown in Figure \ref{fig:safety_considerations}.

\subsection{Experimental Design}
\label{subsec:experimental_design}
Our experimental design is structured to systematically evaluate the effectiveness of LLM-based selective translation for multilingual alignment, addressing the key research questions outlined in the introduction. We compare different translation methodologies, assess the impact of English alignment data, and investigate the benefits of filtering noisy translations.

\begin{figure*}[h]  
    \centering
    \includegraphics[scale=0.1]{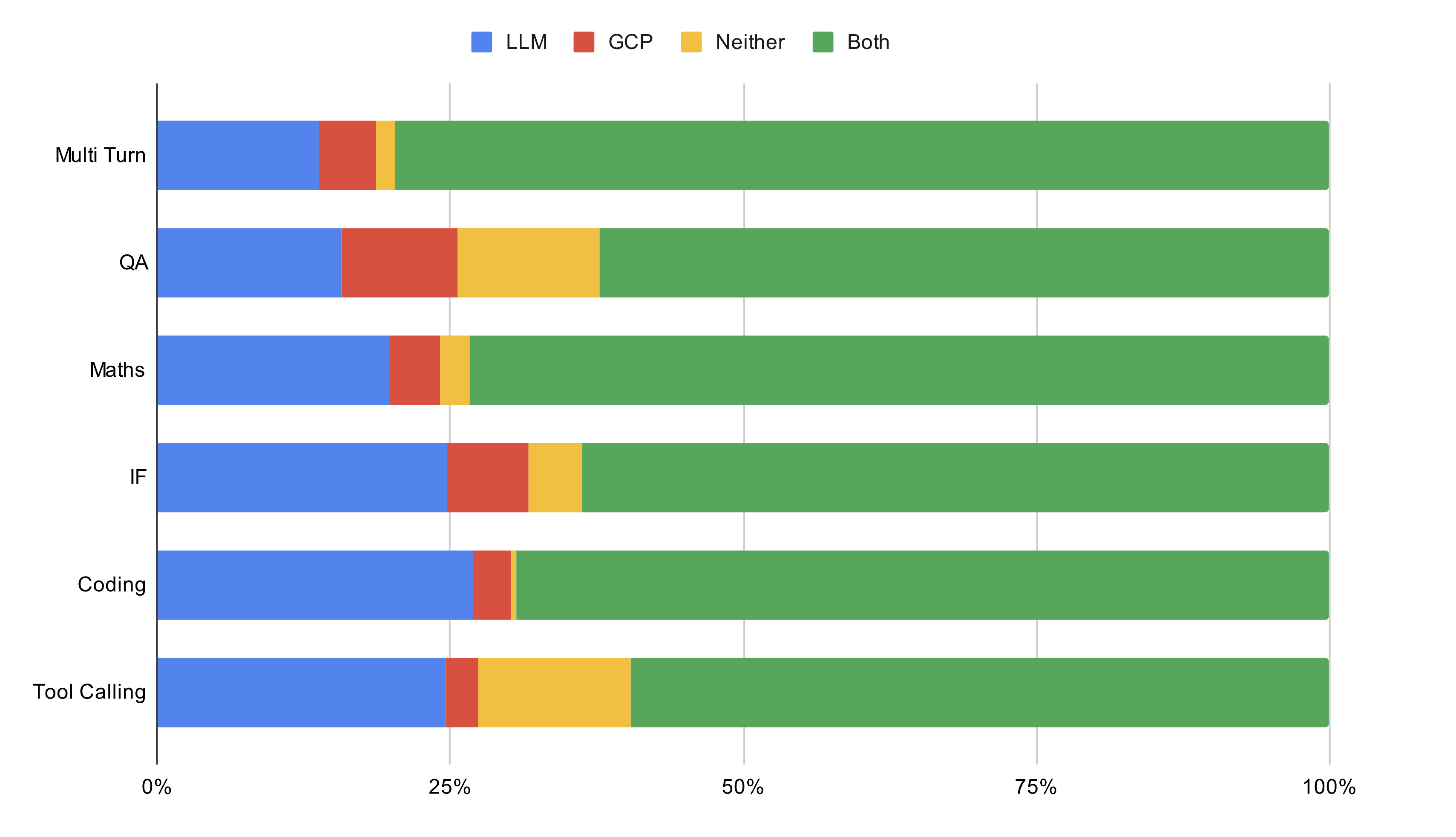}  
    \caption{A/B comparison of translation quality, judged by Llama-3.1-Nemotron-70B-Instruct. The graph illustrates the percentage preference for LLM, GCP, both, or neither across various SFT dataset categories.}
    \label{fig:ab_comparison}
\end{figure*}

\begin{figure}[h]  
    \centering
    \includegraphics[width=\columnwidth]{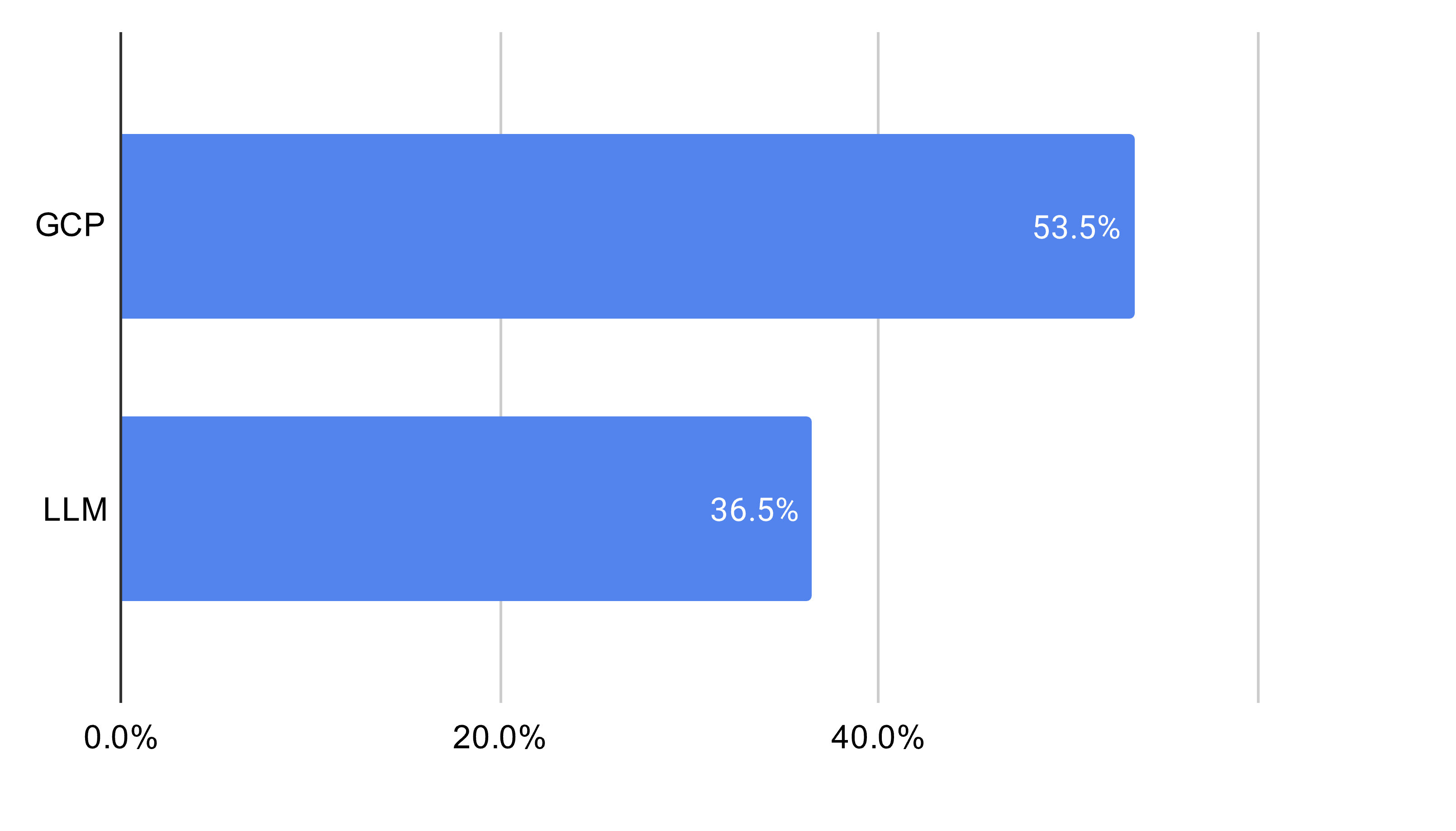}  
    \caption{Percentage of LLM and GCP translated SFT data filtered by the Llama-3.1-Nemotron-70B-Instruct judge model, representing samples not achieving full scores.}
    \label{fig:filtered_data_stats}
\end{figure}

\begin{itemize}
    \item \textbf{GCP vs Llama-3.1-405B Translation}:
This experiment empirically compares the performance of models aligned using data translated by Google Cloud Platform (GCP) against those aligned with data generated via Llama-3.1-405B based selective translation. To achieve this, we perform SFT on the Nemotron-4-Mini-Hindi-4B-Base model. The SFT process utilizes a fixed set of 200k English data samples, combined with varying subsets of translated + filtered Hindi data. The Hindi filtered data subsets consist of 20K, 40K, 60K, 80K, and 100K samples, each randomly selected from a pool of 100K filtered examples. The unfiltered subset comprises 200K samples. Following SFT, the fine-tuned models are benchmarked for their performance across various Hindi test sets, including MTBench, IFEval, and GSM8K, to provide a comprehensive comparison. The Llama-3.1-405B was selected for this study as it represents the largest and highest-quality LLM available for Hindi translations, with human annotators consistently rating its translation quality superior to other competitor models like Llama-3.3-70B-Instruct and Nemotron-4-340B-Instruct. We note that while larger contexts can sometimes lead to "sentence-drop" issues in translation, this is not a concern here as we translate entire prompts and responses, ensuring coherence. Furthermore, larger LLMs like Llama-3.1-405B exhibit greater resilience to the sentence drop issue and a larger context length of 128k.

\item \textbf{Impact of English Alignment Data}:
This experiment aims to assess the necessity of including English data during the SFT phase, or if Hindi data alone is sufficient to achieve the desired performance in the target low-resource language. In this experiment, we perform SFT using only Hindi data. The results from these experiments are then compared against the previous experiments, where both English and Hindi data were incorporated during the SFT process, allowing us to quantify the contribution of English alignment data.

\item \textbf{Impact of Filtering Noisy Translations}:
This experiment investigates the impact of reducing the dataset size through quality filtering on the overall model performance. We compare the performance of the SFT + DPO model with and without applying a filtering step to the training data. Both the SFT and DPO datasets were subjected to this quality filtering process. Post-filtering, the LLM-translated SFT corpus was reduced from its original 200k samples to ~100k samples. Similarly, the LLM-translated DPO corpus also underwent a reduction in size from ~200k to ~100k samples after filtering. The GCP-translated SFT and DPO corpus is reduced to ~90k and ~80k, respectively. The comparison will highlight the benefits of data quality over quantity in multilingual alignment.

\item \textbf{Fluency Analysis}:
To evaluate the fluency of LLM-based selective translation and GCP outputs, the Llama-3.1-Nemotron-70B-Instruct model is employed as an automated evaluator. It is recognized that line-by-line translation, often characteristic of methods like GCP, can lead to inter-sentence disfluencies. Therefore, the assessment specifically targets the naturalness and coherence of the Hindi responses. The Llama-3.1-Nemotron-70B-Instruct model, serving as a Hindi-proficient evaluator, rates responses on a scale of 1-5 across four key criteria: Grammar and Syntax, Fluency and Naturalness, Pacing and Readability, and Cohesion and Coherence. These individual ratings, along with an overall fluency score, are provided, facilitating a quantitative comparison of translation fluency. The prompt used for fluency evaluation is presented in the Appendix \ref{sec:appendix}.
\end{itemize}

\begin{table*}[h!]
\centering
\begin{tabular}{llcccc}
\toprule
\textbf{Training Config} & & \textbf{SubjectiveEval} & \textbf{GSM8K-Hi} & \textbf{IFEval-Hi} & \textbf{MTBench-Hi} \\
\midrule
\multirow{1}{*}{200K En} & -- & 3.71 & 30.10 & 44.17 & 3.44 \\
\midrule
\multirow{2}{*}{200K En + 20K Hi} & LLM & 4.12 & 38.67 & 45.44 & 4.32 \\
                                 & GCP & 4.02 & 36.32 & 43.77 & 4.10 \\
\midrule
\multirow{2}{*}{200K En + 40K Hi} & LLM & 4.29 & 40.79 & 45.92 & 4.67 \\
                                 & GCP & 4.24 & 37.45 & 44.65 & 4.37 \\
\midrule
\multirow{2}{*}{200K En + 60K Hi} & LLM & 4.29 & 42.15 & 45.44 & 4.30 \\
                                 & GCP & 4.13 & 38.36 & 45.04 & 4.26 \\
\midrule
\multirow{2}{*}{200K En + 80K Hi} & LLM & 4.23 & 40.26 & 45.28 & 4.66 \\
                                 & GCP & 3.92 & 39.58 & 45.12 & 4.04 \\
\midrule
\multirow{2}{*}{200K En + 100K Hi} & LLM & 4.15 & 40.86 & 46.39 & 4.62 \\
                                  & GCP & 3.98 & 40.71 & 44.65 & 4.17 \\
\midrule
\multirow{2}{*}{200K En + 200K Hi} & LLM & 4.18 & 43.44 & 43.77 & 4.43 \\
                                  & GCP & 4.05 & 44.50 & 46.63 & 4.55 \\
\bottomrule
\end{tabular}
\caption{Comparison of GCP and Llama-3.1-405B selective translation performance on downstream Hindi tasks. The table details results from SFT models trained on a full English corpus alongside varying percentages of Hindi data. SubjectiveEval is a rating between (1-5), GSM8K-Hi and IFEval-Hi are accuracy (\%), and MTBench-Hi is a rating between (1-10).}
\label{tab:hindi-data-impact}
\end{table*}



\begin{table*}[h!]
\centering
\begin{tabular}{lccccc}
\toprule
\textbf{Training Config} & \textbf{SubjectiveEval} & \textbf{GSM8K-Hi} & \textbf{IFEval-Hi} & \textbf{IFEval-En} \\
\midrule
20K Hi & 4.02 & 29.49 & 36.56 & 34.53 \\
20K Hi + En & 4.12 & 38.67 & 45.44 & 50.84 \\
\midrule
40K Hi & 4.17 & 34.72 & 41.24 & 40.77 \\
40K Hi + En & 4.29 & 40.79 & 45.92 & 50.00 \\
\midrule
60K Hi & 4.21 & 36.09 & 44.09 & 44.00 \\
60K Hi + En & 4.29 & 42.15 & 45.44 & 50.96 \\
\midrule
80K Hi & 4.35 & 35.71 & 45.44 & 46.28 \\
80K Hi + En & 4.23 & 40.26 & 45.28 & 50.96 \\
\midrule
100K Hi & 4.16 & 38.97 & 45.12 & 45.68 \\
100K Hi + En & 4.15 & 40.86 & 46.39 & 50.84 \\
\bottomrule
\end{tabular}%
\caption{The experiments to investigate the impact of training SFT models on either Hindi-only data or a combination of English and Hindi data. Downstream scores are then computed for models trained with different proportions of Hindi content.}
\label{tab:hi-en-impact}
\end{table*}

\begin{table*}[h!]
\centering
\resizebox{\textwidth}{!}{%
\begin{tabular}{llccccc}
\toprule
\textbf{Training Config} & & \textbf{SubjectiveEval} & \textbf{GSM8K-Hi} & \textbf{IFEval-Hi} & \textbf{MTBench-Hi} & \textbf{Fluency} \\
\midrule
\multirow{2}{*}{Filtered SFT - Filtered DPO} 
    & LLM & 4.37 & 43.44 & 55.51 & 4.97 & 4.50 \\
    & GCP & 4.37 & 43.44 & 55.67 & 4.62 & 4.39 \\
    
\midrule
\multirow{2}{*}{Unfiltered SFT - Unfiltered DPO} 
    & LLM & 4.39 & 44.28 & 57.10 & 4.51 & 4.50 \\
    & GCP & 4.24 & 43.59 & 58.84 & 5.01 & 4.42 \\
\bottomrule
\end{tabular}%
}
\caption{Experiments to study the impact of quality filtering on the performance of downstream Hindi tasks. SFT and DPO training were performed using a comprehensive English corpus, in combination with either filtered or unfiltered Hindi translated data. The fluency score is a rating between (1-5).}
\label{tab:combined-sft-dpo}
\end{table*}

\subsection{Evaluation Datasets}
The evaluation of conversational abilities in large language models typically relies on extensive English datasets like IFEval, MTBench, and GSM8K. For Hindi, however, available options such as MILU \cite{verma2024milu}, and Global MMLU \cite{singh2024global} are more limited, primarily focusing on foundational model assessment rather than advanced conversational nuances. Direct translation of English datasets into Hindi often overlooks cultural nuances and linguistic structures, leading to grammatical errors and compounding inherent errors in the translation process. To address this, we adopt a multi-step approach incorporating human oversight to ensure accurate assessment of Hindi language capabilities. The following datasets introduced in \cite{kamath2025benchmarking} were used to benchmark the aligned models trained in this work.
\begin{itemize}
    \item 
\textbf{SubjectiveEval:}
The Hindi SubjectiveEval dataset comprises 91 open-ended questions covering diverse Indian domains, science and technology, mathematics, and thinking ability \cite{joshi2024adapting}. It includes hypothetical scenarios designed to assess analytical reasoning and problem-solving. Model responses are evaluated using an LLM as a judge, specifically GPT-4o, with responses rated on a 1-5 scale.
\item 
\textbf{IFEval-Hi\footnote{\href{https://huggingface.co/datasets/nvidia/IFEval-Hi}{https://huggingface.co/datasets/nvidia/IFEval-Hi}}:}
The Hindi IFEval dataset contains 848 prompts to evaluate the instruction-following ability of LLMs in Hindi. Structured similarly to its English counterpart, it features "verifiable instructions" with heuristically validated responses. These prompts are natively curated by Hindi-proficient specialists to capture local linguistic nuances and Indian cultural context.
\item 
\textbf{GSM8K-Hi\footnote{\href{https://huggingface.co/datasets/nvidia/GSM8K-Hi}{https://huggingface.co/datasets/nvidia/GSM8K-Hi}}:}
Hindi-GSM8K is the GCP-translated version of the English GSM8K test set. Its samples are meticulously reviewed and corrected by human annotators for quality improvement. These problems typically require 2 to 8 steps to solve, primarily involving elementary arithmetic calculations.
\item 
\textbf{MT-Bench-Hi\footnote{\href{https://huggingface.co/datasets/nvidia/MT-Bench-Hi}{https://huggingface.co/datasets/nvidia/MT-Bench-Hi}}:}
The Hindi MTBench dataset consists of 200 multi-turn prompts designed to evaluate the conversational ability of Hindi LLMs. Eighty percent of its samples are natively created by Hindi specialists, with the remaining 20\% translated from the English version, ensuring a balanced and comprehensive evaluation. For this work, we use a subset of 40 samples from MTBench-Hi, focusing on classes such as coding, STEM, math, reasoning, and multiturn interactions. The evaluation is conducted by GPT-4o, with responses rated on a scale of 1-10. It is noted that the scores obtained are on the lower side, as this subset represents areas where Hindi models typically do not excel.
\end{itemize}

\begin{figure*}[!t] 
    \centering 
    \begin{tcolorbox}[
        colback=gray!10, 
        colframe=blue!50, 
        boxrule=0.5pt, 
        arc=2mm, 
        boxsep=5pt, 
        left=5pt, right=5pt, top=5pt, bottom=5pt, 
        width=\textwidth, 
    ]
    \small 
    For LLM alignment in low-resource languages,
    \begin{itemize}[
        nosep, 
        leftmargin=1.5em, 
        label=\color{gray!80}\textbullet, 
    ]
        \item LLM-based selective translation significantly improves model performance.
        \item Mixing translated low-resource data with original English data is crucial for robust alignment.
        \item Filtering translated data for quality is effective and can make training more efficient.
        \item Even small amounts of high-quality translated data offer notable performance gains.
    \end{itemize}
    \end{tcolorbox}
    \caption{Summary of key insights and best practices} 
    \label{fig:key_insights} 
\end{figure*}

\section{Results and Discussion}
This section presents the results of our empirical study comparing LLM-based selective translation and GCP-based regular translation, utilizing Nemotron-4-Mini-Hindi-4B-Base as the base model for all experiments. The base model underwent Supervised Fine-Tuning (SFT) and Direct Preference Optimization (DPO) on various data combinations as detailed in the Section \ref{subsec:experimental_design}. Model performance was evaluated on SubjectiveEval, GSM8K-Hi, IFEval-Hi, and MTBench-Hi datasets. The key findings and best practices are shown in Figure \ref{fig:key_insights}.


\begin{itemize}
    \item 
\textbf{GCP vs Llama-3.1-405B Translation:}
The results of this comparison are presented in Table \ref{tab:hindi-data-impact}. We observe three key findings:
\begin{itemize}
\item 
Models trained on Llama-3.1-405B translations consistently outperform models trained on GCP translations across all benchmark datasets.
\item 
The inclusion of Hindi data alongside English data during training significantly improves performance compared to training on English data alone. Even a small amount, specifically 20k Hindi samples, demonstrates a notable boost in accuracy.
\item 
As the quantity of Hindi data in the SFT datablend increases, the model's accuracy continues to improve, reaching saturation around 60k samples.    
\end{itemize}

\item 
\textbf{Impact of English Alignment Data:}
Table \ref{tab:hi-en-impact} illustrates the impact of incorporating both English and Hindi data during SFT, as opposed to using only Hindi data. While it might seem desirable to align Hindi LLMs solely with the Hindi corpus, our findings indicate that the addition of English data significantly enhances the model's capabilities in mathematics, instruction following, and overall Hindi language proficiency.

\item 
\textbf{Impact of Filtering Noisy Translations:}
Table \ref{tab:combined-sft-dpo} presents the results regarding the impact of filtering noisy translated SFT and DPO data. Approximately 50\% of the translated data was discarded in this process. We observe that models trained on this filtered data perform competitively with those trained on the full, unfiltered dataset. This suggests that filtering can make the training process more efficient by reducing the data volume without significantly compromising accuracy. Furthermore, keeping noisy data does not necessarily degrade performance on downstream tasks.

\item 
\textbf{Translation Quality Analysis:}
The fluency analysis results are detailed in Table \ref{tab:combined-sft-dpo}. We observe that LLM-based selective translations consistently receive higher fluency scores from the LLM-Judge. Figure \ref{fig:ab_comparison} further supports this, showing that a judge LLM (Llama-3.1-Nemotron-70B-Instruct) consistently prefers LLM-based selective translations over GCP translations. This preference is particularly pronounced for instruction-following, coding, and tool-calling samples. Furthermore, Figure \ref{fig:filtered_data_stats} highlights that a greater amount of data is discarded for GCP translations than for LLM, suggesting lower initial quality or adherence to filtering criteria. For comparative results, we make sure that the amount of LLM and GCP translated data is equal. Consequently, for the reported comparative results, we standardized the amount of LLM and GCP translated data.

\end{itemize}
\section{Conclusion}
This study systematically investigated LLM-based selective translation for aligning large language models to low-resource languages, with a specific focus on Hindi. Our experiments consistently demonstrated that this approach significantly enhances model performance compared to traditional GCP translation.

A key finding was the substantial accuracy improvement achieved by incorporating even a small quantity of selectively translated Hindi data. We also found that blending translated Hindi data with original English data is crucial for comprehensive alignment, leading to notable advancements in mathematical reasoning, instruction-following, and general Hindi language proficiency.
The superior fluency and consistent preference for selectively translated outputs, as judged by an LLM-based evaluator, further validate the efficacy of our method. These findings collectively highlight the immense potential of LLM-based selective translation in developing more linguistically inclusive and robust AI systems for low-resource environments.



\section*{Acknowledgements}
This work would not have been possible without contributions from many people at NVIDIA. To mention a few: Yoshi Suhara, Ameya Mahabaleshwarkar, Zijia Chen, Rohit Watve, Oluwatobi Olabiyi, Eileen Long, Mostofa Patwary, and Oleksii Kuchaiev.   

\bibliography{main}

\appendix
\section{Appendix}
\label{sec:appendix}
\begin{figure*}[!htb] 
    \centering 
    \begin{tcolorbox}[
        colback=gray!10,          
        colframe=blue!50,         
        width=\textwidth,         
        arc=5mm,                  
        outer arc=5mm,
        boxsep=5pt,               
        left=10pt, right=10pt,    
        top=10pt, bottom=10pt,    
        boxrule=1pt,              
        coltitle=black,           
        fonttitle=\bfseries\Large 
    ]
    \tiny 
    \begin{verbatim}
You are a Hindi translation assistant. Your task is to translate the following text into Hindi, 
while applying the following rules to determine when to skip translation for specific parts:

- Skip translating the following if they appear in the sentence:
  1. **Programming or coding content** (e.g., code snippets, commands) — retain this exactly as it is.
  2. **URLs, file paths, or email addresses** — leave these unchanged.
  3. **Strongly formatted data** such as tables, lists, or bullet points — maintain their structure and content as is.
  4. **Examples or phrases** where translation would alter their original meaning or usefulness.
  5. **Special characters, mathematical symbols, or technical abbreviations** — do not change these.
  6. **HTML/XML tags or other formatting markers** — keep these intact and unaltered.

As you translate, ensure that the output flows naturally and maintains the overall structure of the sentence. 
Retain non-translatable elements exactly as they are, while translating the rest into Hindi.

Translate the following text:

Text: {{english_text}}

Only return the translated text!
If translation is not needed, return the input text as it-is!

    \end{verbatim}
    \end{tcolorbox}
    \caption{LLM-based selective translation prompt. This is used to translate the entire prompt or response.} 
    \label{fig:prompt_template} 
\end{figure*}

\begin{figure*}[!t] 
    \centering 
    \begin{tcolorbox}[
        colback=gray!10,          
        colframe=blue!50,         
        width=\textwidth,         
        arc=5mm,                  
        outer arc=5mm,
        boxsep=5pt,               
        left=10pt, right=10pt,    
        top=10pt, bottom=10pt,    
        boxrule=1pt,              
        coltitle=black,           
        fonttitle=\bfseries\Large 
    ]
    \tiny 
    \begin{verbatim}
Given the following sentences:

- Source : {{english_text}}
- Target [Hindi]: {{hindi_text}}

Please evaluate the translation using the FAITH metric. For each category, provide a score from 1 to 5 (1 = poor, 5 = excellent). 
Only return the evaluation in the following JSON format:

{
  "Fluency": score, 
  "Accuracy": score, 
  "Idiomaticity": score, 
  "Terminology": score, 
  "Handling_of_Format": score
}

Here are the categories:

1. **Fluency (1-5)**: Does the translation read naturally in the target language, free from grammar or syntax errors?
   - 1: Very poor fluency, difficult to understand.
   - 2: Somewhat fluent but with major grammatical issues.
   - 3: Generally fluent with a few errors.
   - 4: Mostly fluent but may have minor grammatical issues.
   - 5: Perfect grammar, native-like fluency.

2. **Accuracy (1-5)**: How well does the translation preserve the meaning of the source sentence?
   - 1: Meaning significantly changed or lost.
   - 2: Major inaccuracies, important meanings are omitted.
   - 3: Some meaning preserved, but there are notable inaccuracies.
   - 4: Meaning mostly preserved with minor issues.
   - 5: Meaning fully preserved.

3. **Idiomaticity (1-5)**: Are the phrases idiomatic and natural for the target language, 
    fitting its cultural context?
   - 1: Literal translation, very awkward for native speakers.
   - 2: Some idiomatic phrases but mostly awkward.
   - 3: Mixed idiomaticity, some phrases fit while others don't.
   - 4: Mostly idiomatic, with a few non-native phrases.
   - 5: Completely idiomatic and culturally appropriate.

4. **Terminology (1-5)**: Are any specialized terms translated accurately? 
    (If no specialized terms, note as N/A.)
   - 1: Significant errors in terminology.
   - 2: Some incorrect terminology affecting understanding.
   - 3: Mostly correct terminology but with some inconsistencies.
   - 4: All terms correctly translated with minor inconsistencies.
   - 5: All terms correctly and consistently translated.

5. **Handling of Format (1-5)**: Is the formatting (punctuation, capitalization, non-translatable elements) correctly maintained?
   - 1: Significant formatting errors or omissions.
   - 2: Major formatting issues that affect readability.
   - 3: Some formatting errors, but generally readable.
   - 4: Minor formatting issues but mostly preserved.
   - 5: Format fully preserved.

In case there is no translation provided, give -1 to all the categories! If case of non-applicable score, make the score=0

Only return the evaluation JSON! No explanation!
    \end{verbatim}
    \end{tcolorbox}
    \caption{FAITH-based translation quality filtering prompt} 
    \label{fig:prompt_template} 
\end{figure*}

\begin{figure*}[!t] 
    \centering 
    \begin{tcolorbox}[
        colback=gray!10,          
        colframe=blue!50,         
        width=\textwidth,         
        arc=5mm,                  
        outer arc=5mm,
        boxsep=5pt,               
        left=10pt, right=10pt,    
        top=10pt, bottom=10pt,    
        boxrule=1pt,              
        coltitle=black,           
        fonttitle=\bfseries\Large 
    ]
    \tiny 
    \begin{verbatim}
You are an evaluator tasked with assessing the quality of a response to a query using five key metrics: 
Helpfulness, Correctness, Coherence, Complexity, and Verbosity. Provide a score for each metric on a scale of 1-5, 
where 1 indicates poor performance and 5 indicates excellent performance. Then, summarize your reasoning for each score in a brief comment.

Query: {{hindi_prompt}}
Response: {{hindi_response}}

#### Definitions of Metrics and Scoring Guidelines:
- **Helpfulness**: Measures how useful and actionable the response is in addressing the query.
    - 1: Completely unhelpful or irrelevant.
    - 2: Slightly helpful but misses key aspects of the query.
    - 3: Moderately helpful but lacks depth or usability.
    - 4: Mostly helpful with minor gaps in utility.
    - 5: Extremely helpful, fully addressing the query with clear, actionable information.

- **Correctness**: Evaluates whether the response is factually accurate and free of errors.
    - 1: Contains major factual inaccuracies or misleading information.
    - 2: Includes some accurate information but has notable errors.
    - 3: Mostly accurate but with minor errors or omissions.
    - 4: Accurate with negligible issues.
    - 5: Completely accurate and reliable.

- **Coherence**: Assesses whether the response is logically structured and easy to follow.
    - 1: Illogical, disorganized, or hard to understand.
    - 2: Poorly structured with noticeable issues in logical flow.
    - 3: Somewhat coherent but with occasional disorganization.
    - 4: Mostly coherent and well-organized with minor issues.
    - 5: Perfectly coherent, logically structured, and easy to follow.

- **Complexity**: Measures whether the response appropriately balances depth and complexity for the query.
    - 1: Overly simplistic or excessively complicated without justification.
    - 2: Either too simple or too complex, with limited balance.
    - 3: Moderately balanced but could improve in complexity or simplicity.
    - 4: Mostly balanced, with only minor adjustments needed.
    - 5: Perfectly balanced, with the right level of complexity for the query.

- **Verbosity**: Evaluates whether the response is concise and avoids unnecessary elaboration.
    - 1: Excessively verbose or overly terse, failing to strike a balance.
    - 2: Somewhat verbose or overly brief with noticeable issues.
    - 3: Moderately concise but could improve in eliminating redundancy or brevity.
    - 4: Mostly concise with minor verbosity or brevity issues.
    - 5: Perfectly concise, providing just the right amount of information.

#### Output Format:
Provide the evaluation in the following JSON format:
{
    "Helpfulness": score,
    "Correctness": score,
    "Coherence": score,
    "Complexity": score,
    "Verbosity": score
}

In case there is no translation provided, give -1 to all the categories!
If case of non-applicable score, make the score=0

Only return the evaluation JSON! No explanation!
    \end{verbatim}
    \end{tcolorbox}
    \caption{Alignment-based quality filtering prompt} 
    \label{fig:prompt_template} 
\end{figure*}

\begin{figure*}[!t] 
    \centering 
    \begin{tcolorbox}[
        colback=gray!10,          
        colframe=blue!50,         
        width=\textwidth,         
        arc=5mm,                  
        outer arc=5mm,
        boxsep=5pt,               
        left=10pt, right=10pt,    
        top=10pt, bottom=10pt,    
        boxrule=1pt,              
        coltitle=black,           
        fonttitle=\bfseries\Large 
    ]
    \tiny 
    \begin{verbatim}
You are a helpful Evaluator. Your task is to critically assess the fluency of responses given by a model to user questions in Hindi. 
        
You will be presented with a chat containing user question and bot response pairs in Hindi. 
Your goal is to evaluate the fluency of the response on a scale of 1-5, with 1 being the lowest and 5 being the highest. 
You are proficient in the Hindi language, so you should consider the nuances and context of the language in your evaluation. 
Your evaluation should be based on the following criteria:

1. Grammar and Syntax: Is the response grammatically correct and properly structured in Hindi?
2. Fluency and Naturalness: Does the response sound natural and fluent, as if it were written or spoken by a native Hindi speaker?
3. Pacing and Readability: Is the response paced well and easy to read or understand for a Hindi-speaking audience?
4. Cohesion and Coherence: Are the ideas logically connected, and does the response flow smoothly?

You will rate each criterion individually and then provide an overall fluency rating from 1 to 5.

Here is the chat:

User Question:  
{hindi_prompt}

Bot Response:  
{hindi_response}

At the end, provide the ratings in a JSON format with appropriate keys and values.

Example JSON format:  
"grammar_and_syntax": 4,
"fluency_and_naturalness": 5,
"pacing_and_readability": 4,
"cohesion_and_coherence": 5,
"overall": 4

Return the JSON object with the above 5 parameters, with all ratings as integers. 
Do not include anything else.
    \end{verbatim}
    \end{tcolorbox}
    \caption{Fluency evaluation prompt} 
    \label{fig:prompt_template} 
\end{figure*}


\end{document}